\documentclass[conference]{IEEEtran}

\usepackage{times}
\usepackage{helvet}
\usepackage{courier}
\usepackage{textcomp}

\usepackage{psfig}
\usepackage{subfig}
\usepackage[linesnumbered,ruled,vlined]{algorithm2e}
\usepackage{floatflt}
\usepackage{rotating}
\usepackage{multicol}
\usepackage{multirow}
\usepackage{color}
\usepackage{epsfig}
\usepackage{amsmath}
\usepackage{amsfonts}
\usepackage{array}
\usepackage{flushend}

\begin{document}
%
\title{Deep Learning Towards Mobile Applications}


\author{\IEEEauthorblockN{Ji Wang\IEEEauthorrefmark{1},
Bokai Cao\IEEEauthorrefmark{2},
Philip S. Yu\IEEEauthorrefmark{2}\IEEEauthorrefmark{3},
Lichao Sun\IEEEauthorrefmark{2},
Weidong Bao\IEEEauthorrefmark{1}, and
Xiaomin Zhu\IEEEauthorrefmark{1}}
\IEEEauthorblockA{\IEEEauthorrefmark{1}College of Systems Engineering,
National University of Defense Technology, Changsha, Hunan, P. R. China}
\IEEEauthorblockA{\IEEEauthorrefmark{2}Department of Computer Science, University of Illinois at Chicago, Chicago,
IL, USA}
\IEEEauthorblockA{\IEEEauthorrefmark{3}Institute for Data Science, Tsinghua University, Beijing, P. R. China}
}

\maketitle

\begin{abstract}
Recent years have witnessed an explosive growth of mobile devices. Mobile devices are permeating every aspect of our daily lives. With the increasing usage of mobile devices and intelligent applications, there is a soaring demand for mobile applications with machine learning services. Inspired by the tremendous success achieved by deep learning in many machine learning tasks, it becomes a natural trend to push deep learning towards mobile applications. However, there exist many challenges to realize deep learning in mobile applications, including the contradiction between the miniature nature of mobile devices and the resource requirement of deep neural networks, the privacy and security concerns about individuals' data, and so on. To resolve these challenges, during the past few years, great leaps have been made in this area. In this paper, we provide an overview of the current challenges and representative achievements about pushing deep learning on mobile devices from three aspects: training with mobile data, efficient inference on mobile devices, and applications of mobile deep learning. The former two aspects cover the primary tasks of deep learning. Then, we go through our two recent applications that apply the data collected by mobile devices to inferring mood disturbance and user identification. Finally, we conclude this paper with the discussion of the future of this area.
\end{abstract}


%
\IEEEpeerreviewmaketitle

\section{Introduction}
The past few years have witnessed an explosive growth of mobile devices which is expected to continue in the next decades. It is predicted that mobile devices will reach 5.6 billion, accounting for 21\% of all networked devices in 2020 \cite{Cisco2016}. By the end of 2023, more than 90\% adults in developed countries will own at least one mobile device. Mobile devices, permeating almost every aspect of our daily lives, are redefining how people live and interact with each other.

Meanwhile, machine learning (ML) has become commonly used in mobile applications like object recognition, language translation, health monitoring, malware detection \cite{He2017,cao2017deepmood,Li2017modeling,sun2016sigpid,li2016droidclassifier} etc. Due to the frequent interaction between mobile devices and users in daily lives, mobile devices collect a large amount of data concerning users' behavior, preference and habits, which makes them promising resources for machine learning applications. According to the report from Delloitte \cite{Delloitte2017}, the penetration of ML-featured applications among mobile application used by adult users in developed countries surpasses 60\%. Machine learning will become a core element of future mobile devices.

Recently, the most exciting advancement of machine learning mainly comes from the field of deep learning (DL) whose unprecedented performance has beaten many records achieved by the traditional machine learning algorithms. Deep learning has revolutionized how the world processes, models, and interprets data \cite{Goodfellow2016DL}. Inspired by the outstanding performance of deep learning, people naturally attempt to push deep learning on mobile devices to provide high-quality intelligent services. It is believed that deep learning will play a paramountly important role in the evolution of mobile applications \cite{Lane2017}. Despite the attractive prospect, the current research of blending deep learning and mobile devices is just the beginning. Deep learning based applications can be broken down to two primary tasks, i.e., training and inference. In the training phase, a large set of training data are used to adjust the trainable parameters in the deep neural network (DNN) automatically by the gradient descent algorithms \cite{Kingm2015,Duchi2011,Rmsprop}. Given a DNN model trained for a specific application, we are able to apply it to the inference task such as recognizing an image it has never seen. Due to the limited computation capacity and battery capacity, neither of these two tasks is trivial for mobile devices.

Training a deep neural network containing hundreds of millions of parameters can easily introduce huge resource demands that are far beyond the capacity of mobile devices. Hence, rather than training a DNN model on mobile devices, the current research mainly focuses on how to make full utilization of distributed data generated by individuals' mobile devices with the consideration of safety, confidentiality, and privacy. Training is not the only challenging process in pushing deep learning on mobile devices. Even the inference step using a DNN model on a mobile device can be difficult \cite{Delloitte2017}. The large DNN models exceed the limited on-chip memory of mobile devices, and thus they have to be accommodated by the off-chip memory, which unfortunately consumes significantly more energy \cite{Han2015learning,Ding2017}. What is worse, the enormous dot products aggravate the burden of processing units. Running a deep learning application can easily dominate the whole system energy consumption. To resolve these challenges, both the academia and the industry have studied how to execute DNN efficiently on mobile devices, which is critical to push deep learning on mobile devices.

In this paper, we provide an overview of the current challenges and achievements about pushing deep learning on mobile devices. The discussion mainly focuses on how to utilize the data generated by individuals' mobile devices and how to execute DNNs on mobile devices with high efficiency. In addition, we will introduce our latest works about deep learning on mobile devices, especially the applications of using the data generated by the mobile devices.

\section{Utilization of Mobile Data}
In this section, we introduce the works about how to utilize the data generated by individuals' mobile devices to train a mobile deep learning model. Firstly, the distributed training, the foundation of using distributed data to train DNNs, is detailed. Then, we introduce the latest training scheme, federated learning, proposed by Google. Finally, we discuss the privacy concerns and the corresponding privacy-preserving methods.

\subsection{Distributed Training}
Mobile devices generate and record a large volume of data about their users. To provide an intelligent mobile service, it is in high demand to create DNNs by using the data on individuals' mobile devices. Due to the resource-hungry nature of training phase, it is infeasible to train DNNs on resource-constrained mobile devices. Training DNNs still relies on cloud data centers and high-performance computers. Nonetheless, the privacy risk and violation make it prohibited to transfer individuals' data directly to third parties. In order to utilize these data without exposing them, it is needed to propose a distributed training scheme.

Adjusting the trainable parameters in a DNN is the core process of training. The gradient descent algorithm \cite{Avriel2003} and its variants are the most commonly used algorithms for training DNNs. Therefore, the main problem of distributed training is to design a distributed gradient descent algorithm. Shokri and Shmatikov designed a distributed selective stochastic gradient (SGD) algorithm which realizes distributed and collaborative training by using data from different sources \cite{Shokri2015}.

In the proposed distributed selective SGD, the participants who contribute their own data train the DNN independently and concurrently. After each iteration, the participants upload the gradients of some selective parameters to the global parameter server. The global parameter server maintains the global parameters that are determined by the sum of all gradients gathered from participants. Each participant downloads some global parameters to update its local model. The participants can create a local DNN model by learning more from other participants' data.

\begin{figure}[htb]
\centering
\includegraphics[width=3.2in]{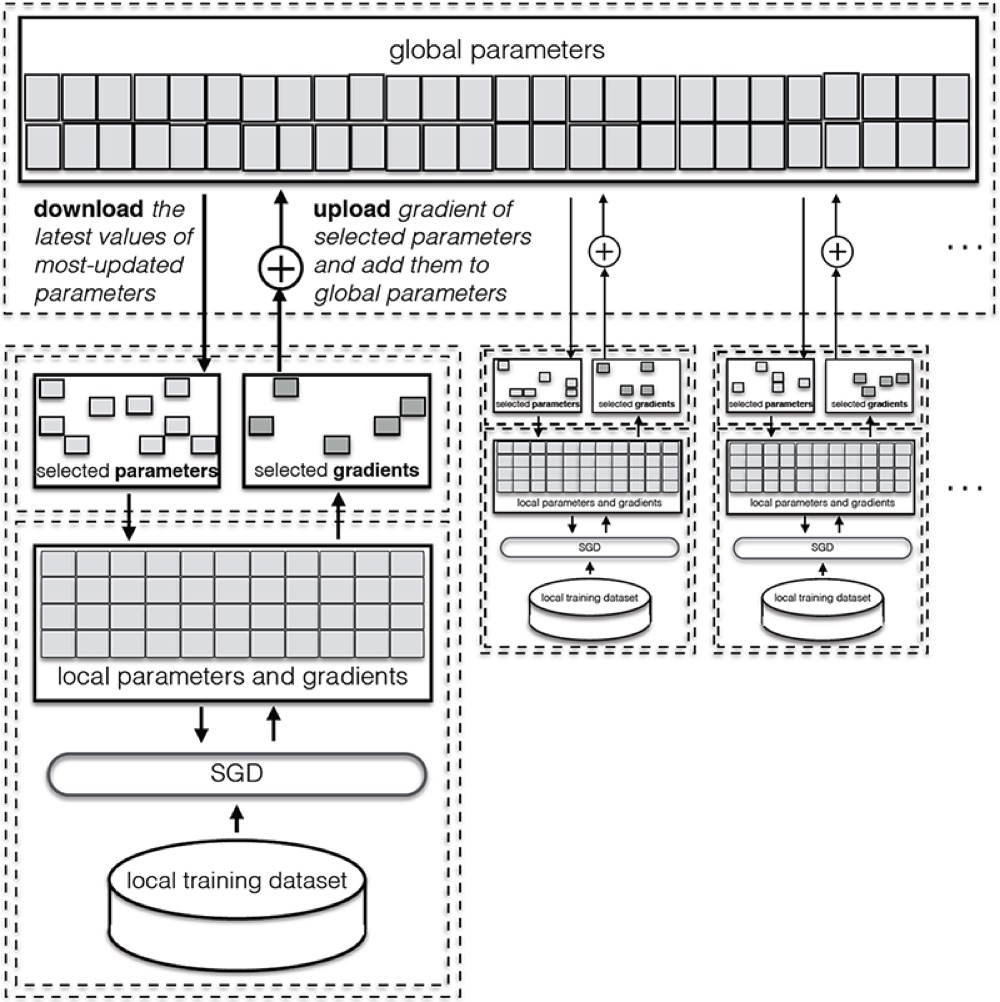}
\caption{Framework of distributed selective SGD \cite{Shokri2015}.} \label{fig:dis}
\end{figure}

Fig. \ref{fig:dis} shows the framework of distributed selective SGD. $N$ participants who have local datasets participate in the distrusted training. Each participant uses its local dataset to train their local DNN model by the standard SGD. No interaction with other participants or the global parameter server is needed during the local training. The gradients of selected parameters are then uploaded to the global parameter server to update the global parameters. The global parameter server maintains the parameters of DNN model based on the gradients uploaded by participants. The participants can download the most recent parameters from the global parameter server to update their local DNN models. This scheme realizes the distributed training for DNN models without explicitly sharing of different participants' local data, which is the foundation of utilizing data from different individuals' mobile devices.

\subsection{Federated Training on Mobile Devices}
In order to train a model from the data generated and captured by mobile devices, Google introduced a new training scheme called federated training \cite{McMahan2017}. Federated training makes it possible to train a shared DNN model while keeping all the individual's data local. Each participant of federated training, i.e., one mobile device, downloads the shared model from the cloud, improves it by learning from local data, and then summarizes the changes to the cloud to improve the shared model. Considering the limited bandwidth and the high energy cost of wireless communication, federated training tries to fully utilize processors in mobile devices to generate updates with higher quality than simple gradient steps in distributed selective SGD.

Suppose there are $K$ participants in the federated training who share a common DNN model with the loss function $ \mathcal{L}$. For participant $k$, its gradient $g_k = \nabla \mathcal{L}_k(\omega _t)$, where $\omega _t$ is the parameters of the current DNN model. In the traditional distributed SGD such as the distributed selective SGD, the global parameter server updates the parameters as:
\begin{equation*}
{\omega _{t + 1}} \leftarrow {\omega _t} - \eta \sum\limits_{k = 1}^K {\frac{{{n_k}}}{n}{g_k}},
\end{equation*}
where $\eta$ is the learning rate, $n_k$ is the size of local dataset, and $n$ is the sum of the size of all datasets. Then, the equivalent form can be given as follows:
\begin{equation*}
\begin{array}{l}
\forall k,\omega _{t + 1}^k \leftarrow {\omega _t} - \eta {g_k},\\
{\omega _{t + 1}} \leftarrow \sum\limits_{k = 1}^K {\frac{{{n_k}}}{n}\omega _{t + 1}^k} .
\end{array}
\end{equation*}

It means that the participant computes one step of gradient descent on the current parameters, and the global parameter server aggregates these gradients uploaded by participants to update the current parameters. Uploading gradients after one step of gradient descent indicates more round of iterations before the convergence. It is natural to let the participant perform $\omega _{t + 1}^k \leftarrow {\omega _t} - \eta {g_k}$ multiple times before the uploading to generate gradients with higher quality. The experimental results from \cite{McMahan2016} show that the improved scheme is able to use 10-100x less communication compared to a naively distributed SGD.

Although the federated training achieves higher efficiency than the naively distributed SGD, training a DNN model on mobile devices is still a tough work which requires sophisticated scheduling. Google pointed out that training happens only when the mobile device is idle, plugged in, and on a free wireless connection to avoid possible adversarial impact on the mobile device's performance.

\subsection{Privacy-Preserving Training}
The above distributed training methods enable mobile devices to learn a shared DNN model without the explicit exposion of individual's data. Nonetheless, the gradients uploaded by participants may still reveal the features of local training data, which makes it susceptible to powerful attacks such as a generative adversarial network based method proposed in \cite{Hitaj2017}. In order to protect the participants' privacy, a series of works have been proposed \cite{Shokri2015,Abadi2016,Papernot2017,McMahan2017learning}, among which the differential privacy plays an important role.

Differential privacy is a concept of privacy tailored to the privacy-preserving data analysis. It aims at providing provable privacy guarantee for sensitive data and is increasingly regarded as a standard notion for rigorous privacy \cite{Beimel2014}. An algorithm is differentially private when the probability of generating a particular output is not affected very much by whether one data item is in the input \cite{Dwork2011diff}. A well-designed differentially private algorithm can provide aggregated representations about a set of data items without leaking information of any data item.

The property of differential privacy theory makes it a foundation to design privacy-preserving training approaches. Shokri et al. \cite{Shokri2015} presented a privacy-preserving distributed SGD to enable multiple data owners to collaborate on training a DNN, in which the sparse vector technique was introduced to provide differential privacy. Abadi et al. \cite{Abadi2016} designed a new differential privacy mechanism called moments accountant in SGD to reduce the privacy budget. Papernot et al. \cite{Papernot2017} designed a generally applicable framework, private aggregation of teacher ensembles, to guarantee the privacy of sensitive training data. It trained a student model to predict an output chosen by noisy voting among all of the teacher models which are trained by the sensitive data locally. The individual teacher model and its parameters are inaccessible to control the privacy budget. Based on the federated training, the researchers from Google use the differential privacy theory to control the participants' privacy loss \cite{McMahan2017learning}. Several modifications to the non-private federated training are made to achieve differential privacy:

\begin{itemize}
  \item Rather than always selecting the fixed participants, the participants are selected independently with probability $p$;
  \item The updates generated by participants are bounded by a specific L2 norm;
  \item An estimator for weighted aggregation is introduced so that the moments accountant \cite{Abadi2016} can be applied;
  \item Sufficient Gaussian noise is added to the final average update.
\end{itemize} 

Through these modifications, the federated training becomes a user-level differentially private training approach for DNN models. The empirical results on a realistic dataset show that it can guarantee the differential privacy without losing accuracy of DNN models. It is feasible to use data from individuals' mobile devices to train a DNN model with privacy guarantees for real-world applications.

\section{Efficient Inference on Mobile Devices}
Compared with the training phase, the inference phase requires much fewer resources, which makes it possible to run the inference on mobile devices locally. Despite this possibility, running inference is still an energy-consuming task for mobile devices. Therefore, there are two choices to run a trained DNN model: inference on the cloud server and inference on the local mobile device.

\begin{figure}[htb]
\centering
\includegraphics[width=3.2in]{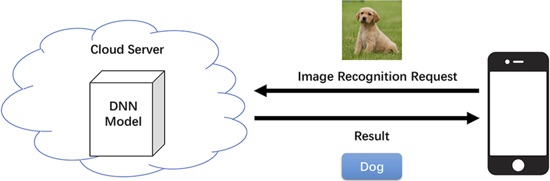}
\caption{Illustration of inference on the cloud server.} \label{fig:remote}
\end{figure}

Fig. \ref{fig:remote} demonstrates how inference on the cloud server works. The trained DNN model is deployed on the cloud server and publishes an interface for interacting with users' mobile app over the internet. Take an image recognition application as an example. The mobile device sends the image to the cloud server over the internet. The cloud server receives the image and feeds it into the DNN model for inference after which the cloud server sends the results back to the mobile device. Inference on the cloud server deploys the complex DNN model on the cloud server and keeps the mobile app simple. All the resource-hungry tasks are accomplished by the cloud server. Besides, the DNN model can be updated anytime without the modification of local apps. For deep learning service providers, they are free from the worry about the leakage of valuable and highly tuned DNNs as the trained model are deployed on the server under control. Nonetheless, inference on the cloud server has two serious problems. Firstly, it requires the internet access to enable the interaction between the mobile device and the cloud server, which sometimes is infeasible in reality. Secondly, submitting individual's data to the cloud server may violate the privacy requirement. Many researchers have made efforts to resolve these two problems. Teerapittayanon et al. \cite{Teerapittayanon2017} designed a distributed DNN architecture across the cloud, the edge, and the mobile devices, which allowed the combination of fast and localized inference on mobile devices and complex inference in cloud servers. For the privacy concern, Osia et al. \cite{Osia2017} designed a hybrid deep learning architecture for private inference across mobile devices and clouds. The Siamese network was used to protect against undesired inference attacks, which in essence provided $k$-anonymity protection. Li et al. \cite{Li2017} proposed a flexible framework for private deep learning. Before revealing data to clouds, it was transformed by the local neural network whose topological structure, including the number of layers, the depth of output channels, and the subset of selected channels, was variable to protect against the reconstruction attacks. 

For inference on the local mobile device, the trained DNN model is deployed on the local mobile device. There is no interaction with other devices during the inference. One of the benefits of doing inference on a mobile device is that users can use the deep learning applications even without the internet access. In addition, as the data would not be transmitted outside the mobile device, the individuals' privacy can be guaranteed. However, this scheme also suffers from a series of drawbacks. Firstly, the size of the app combined with the DNN model may be extraordinarily large, which makes it difficult to install and update this app. Secondly, doing inference of complex DNN models requires a lot of computation resources, which can quickly drain the limited energy of mobile devices. In the past few years, several methods have been proposed to compress DNN models and prune the computation during inference. Han et al. \cite{Han2016} took three steps to compress the DNN model. Firstly, the network was pruned by learning only the important connection. Then, they quantized the parameters to enforce parameter sharing. Finally, the Huffman coding was applied. Ding et al. \cite{Ding2017} used the fast fourier transform based multiplication to reduce the computational complexity and the storage cost simultaneously. Howard et al. \cite{Howard2017} proposed a group of models named MobileNets which are based on the streamlined architecture for vision applications on mobile devices. 

We will introduce one of our latest researches about cloud-based inference with the consideration of privacy. Then, we will briefly discuss the researches about compressing DNN models.

\subsection{Private Cloud-Based Inference}
\begin{figure}[tb]
\centering
\includegraphics[width=3.3in]{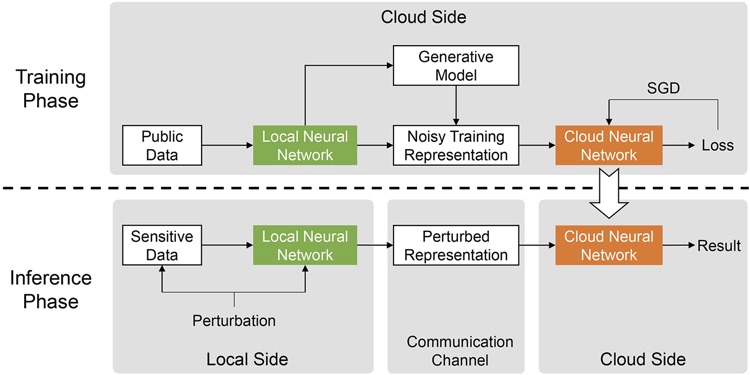}
\caption{Framework of cloud-based solution.} \label{fig:framework}
\end{figure}

In order to relieve the heavy burden of doing inference on mobile devices, we propose a cloud-based solution where the shallow portions of a DNN are deployed on mobile devices while the complex and large parts are offloaded to the cloud server \cite{wang2018not}. A fast transformation is first carried out on the input data locally. Then, the transformed data are revealed to the cloud server for the time and energy consuming inferences. But this solution presents obvious privacy issues with transmitting data from mobile devices to the cloud server. Once an individual reveals its data consisting of sensitive information to cloud server for further inferences, it is almost impossible for the individual to control or influence the usage of the data. Hence, it is a critical requirement to lessen the privacy risk when implementing the cloud-based solution. To this end, we designed a framework as shown in Fig. \ref{fig:framework}.

The cloud-based solution relies on the mobile cloud environment and divides the DNN into the local-side part and the cloud-side part. The local neural network is derived from the pretrained DNN whose structure and weights are frozen. The cloud neural network is fine-tuned in the training phase. The whole training phase and the complex inference phase are performed in the cloud server. Mobile devices merely undertake the simple and light-weight feature extraction and perturbation.

In the inference phase, the sensitive data is transformed by the local neural network to extract the general features embedded in it. For preserving privacy, the transformation is perturbed by both nullification and random noise which satisfy the differentially private requirements. Then, the perturbed representations are transmitted to the cloud for further complex inference. Because the data to be transmitted are the abstract representation of the raw data, the size of the data to be transmitted is smaller than that of the raw data. The local initial transformation can reduce communication cost compared with transmitting raw data directly.

In the training phase, we use the public data of the same type as the sensitive data to train the cloud neural network. In order to improve the robustness of the cloud neural network to the noise, we propose a noisy training method where both raw training data and generative training data are fed into the network to tune the weights. The key component in noisy training is the generative model that generates sophisticated noisy training samples based on the public data. In addition, it is worth noting that the training phase and the inference phase can run in parallel once we get an initial cloud neural network. Benefiting from the transfer learning, we only need to train the cloud neural network for different datasets and tasks while keeping the local neural network frozen. It indicates that all the works in the cloud side are transparent to the end mobile devices. The cloud neural network can be upgraded online without the service interruption to end users. This transparency property facilitates the deep learning service on one hand and on the other hand protects the intellectual property of deep learning service providers.

In this framework, the differential privacy is guaranteed by the nullification and random noise added in the local transformation. To alleviate the adverse impact of random noise on the DNN model's performance, we propose a novel training method called noisy training where deliberate noisy samples are injected into the training data. The preliminary experimental results show that this solution can not only preserve users’ privacy but also improve the inference performance.

\subsection{Model Compression and Acceleration}
Model compression and acceleration is another efficient method to reduce the DNN model's storage and computational cost. By compressing the large DNN model, the resource and energy requirement of inference on mobile devices can be significantly relieved. Currently, there are three main approaches for DNN model compression \cite{Cheng2017}: (1) parameter pruning and sharing; (2) low-rank factorization; (3) model distillation.

The intuition of parameter pruning and sharing is that many parameters in the network are not crucial to the performance of the model. There are three techniques to prune the parameters, i.e., network quantization, weight and connection pruning, and structural matrix. Network quantization compress the DNN by reducing the bits required to depict the parameters in the network \cite{Gong2014, Wu2016, Gupta2015}. Weight and connection pruning tries to prune the redundant weights in the DNN model. The core idea of structural matrix is to describe an $m \times n$ matrix by using a structured matrix with much fewer parameters than $mn$ \cite{Cheng2015an}.

The low-rank factorization is used to reduce the convolution layer whose kernel can be regarded as a 4D tensor. A 4D tensor usually has a large amount of redundancy which can be removed by the low-rank factorization \cite{Denton2014}. It should be noted that the fully-connected layer can be considered as a 2D matrix so the low-rank factorization can also be employed.

The model distillation compresses the DNNs into shallower ones by mimicking the function of the original complex DNN. This approach can be viewed as transferring knowledge from a large teacher model into a small student model \cite{Hinton2015}.

\section{Applications of Mobile Deep Learning}
Two applications using the data generated by mobile devices will be introduced in this section. The two applications apply the data collected by mobile devices to inferring mood disturbance and user identification, respectively.

\subsection{Mood Disturbance Inference}
The wide use of mobile devices presents new opportunities in the study and treatment of psychiatric illness including the ability to study the manifestations of psychiatric illness in the setting of patients' daily lives in an unobtrusive manner and at a level of detail that was not previously possible. Continuous real-time monitoring in naturalistic settings and collection of automatically generated mobile device data that reflect illness activity could facilitate early intervention and have a potential use as objective outcome measures in efficacy trials \cite{ankers2009objective,bopp2010longitudinal,faurholt2016behavioral}.

The data used in this work were collected from the BiAffect\footnote{http://www.biaffect.com/} study. During a preliminary data collection phase, for a period of 8 weeks, 40 individuals were provided a Galaxy Note 4 mobile phone which they were instructed to use as their primary phone during the study. This phone was loaded with a custom keyboard that replaced the standard Android OS keyboard. The keyboard collected metadata consisting of keypress entry time and accelerometer movement and uploaded them to the study server. Three types of metadata collected are detailed as follows:

\begin{itemize}
  \item Alphanumeric characters. Due to privacy reasons, we only collected metadata for keypresses on alphanumeric characters, including duration of a keypress, time since last keypress, and distance from last key along two axisex.
  \item Special characters. We use one-hot-encoding for typing behaviors other than alphanumeric characters, including auto-correct, backspace, space, suggestion, switching-keyboard and other. They are usually sparser than alphanumeric characters.
  \item Accelerometer values. Accelerometer values are recored every 60ms in the background during an active session regardless of a person's typingspeed, thereby making them much denser than alphanumeric characters.
\end{itemize}

We study these dynamics metadata on a session-level. A session is defined as beginning with a keypress which occurs after 5 or more seconds have elapsed since the last keypress and continuing until 5 or more seconds elapse between keypresses. The duration of a session is typically less than one minute. In this manner, each participant would contribute many samples, one per phone usage session, which could benefit data analysis and model training.
\begin{figure}[tb]
\centering
\includegraphics[width=3.3in]{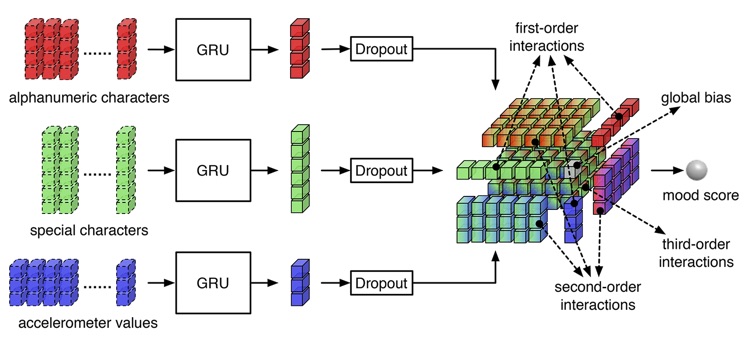}
\caption{The architecture of DeepMood \cite{cao2017deepmood}.} \label{fig:architecture}
\end{figure}

In \cite{cao2017deepmood}, a deep architecture, named DeepMood, is proposed to infer mood disturbance and severity from mobile phone typing dynamics. The architecture is illustrated in Fig.~\ref{fig:architecture}. In particular, it is an end-to-end late fusion approach for modeling the multi-view time series data. In the first stage, each view of the time series is separately modeled using Gated Recurrent Unit (GRU) \cite{cho2014learning} which is a simplified version of Long Short-Term Memory (LSTM) \cite{hochreiter1997long}. A typical GRU keeps hidden states over a sequence of elements and updates the hidden state $\mathbf{h}_k$ by the current input $\mathbf{x}_k$ as well as the previous hidden state $\mathbf{h}_{k-1}$ with a recurrent function as follows:
\begin{equation}
\begin{aligned}
\mathbf{r}_k &= \text{sigmoid} (\mathbf{W}_r\mathbf{x}_k + \mathbf{U}_r\mathbf{h}_{k - 1}) \hfill \\
\mathbf{z}_k &= \text{sigmoid} (\mathbf{W}_z\mathbf{x}_k + \mathbf{U}_z\mathbf{h}_{k - 1}) \hfill \\
\tilde{\mathbf{h}}_k &= \text{tanh} (\mathbf{W}\mathbf{x}_k + \mathbf{U}(\mathbf{r}_k \odot \mathbf{h}_{k - 1})) \hfill \\
\mathbf{h}_k &= \mathbf{z}_k \odot \mathbf{h}_{k - 1} + (1 - \mathbf{z}_k) \odot \tilde {\mathbf{h}}_k \hfill \\
\end{aligned}
\end{equation}
where $\odot$ is the element-wise multiplication operator, a reset gate $\mathbf{r}_k$ allows the GRU to forget the previously computed state $\mathbf{h}_{k-1}$, and an update gate $\mathbf{z}_k$ balances between the previous state $\mathbf{h}_{k-1}$ and the candidate state $\tilde{\mathbf{h}}_k$. The hidden state $\mathbf{h}_k$ can be considered as a compact representation of the input sequence from $\mathbf{x}_1$ to $\mathbf{x}_k$.

The multi-view time series data are then fused in the second stage by exploring interactions across the output vectors from each view, where three alternative approaches are developed following the idea of Multi-view Machines \cite{cao2016multi}, Factorization Machines \cite{rendle2012factorization}, or in a fully connected fashion. Let us denote the output vectors at the end of a sequence from the $p$-th view as $\mathbf{h}^{(p)}$. In this manner, $\{\mathbf{h}^{(p)}\in\mathbb{R}^{d_h}\}_{p=1}^m$ can be considered as multi-view data where $m$ is the number of views.

\noindent\textbf{Fully connected layer}.
In order to generate a prediction on the mood score, a straightforward idea is to first concatenate features from multiple views together, {\em i.e.}, $\mathbf{h} = [\mathbf{h}^{(1)}; \mathbf{h}^{(2)}; \cdots; \mathbf{h}^{(m)}] \in \mathbb{R}^d$, where $d$ is the total number of multi-view features, and typically $d=md_h$ for one-directional GRU and $d=2md_h$ for bidirectional GRU. The concatenated hidden states $\mathbf{h}$ are then fed into one or several fully connected neural network layers with a nonlinear function $\sigma(\cdot)$ in between \cite{cao2017deepmood}.
\begin{equation}
\begin{aligned}
\mathbf{q} &= \text{relu}(\mathbf{W}^{(1)}[\mathbf{h}; 1]) \\
\hat{\mathbf{y}} &= \mathbf{W}^{(2)}\mathbf{q}
\end{aligned}
\end{equation}
where $\mathbf{W}^{(1)} \in \mathbb{R}^{k' \times (d+1)}, \mathbf{W}^{(2)} \in \mathbb{R}^{c \times k'}$, $k'$ is the number of hidden units, $c$ is the number of classes, and the constant signal ``1'' is to model the global bias.

\noindent\textbf{Factorization Machine layer}.
Rather than capturing nonlinearity through the transformation function, one can explicitly model feature interactions between input units \cite{cao2017deepmood}.
\begin{equation}
\begin{aligned}
\mathbf{q}_a &= \mathbf{U}_a\mathbf{h} \\
b_a &= \mathbf{w}_a^T[\mathbf{h}; 1] \\
\hat{y}_a &= \text{sum}([\mathbf{q}_a\odot\mathbf{q}_a;b_a])
\end{aligned}
\label{eq:fm}
\end{equation}
where $\mathbf{U}_a \in \mathbb{R}^{k \times d}, \mathbf{w}_a \in \mathbb{R}^{d+1}$, $k$ is the number of factor units, and $a$ denotes the $a$-th class. 

\noindent\textbf{Multi-view Machine layer}.
In contrast to modeling up to the second-order feature interactions between all input units as in the Factorization Machine layer, one can further explore all feature interactions up to the $m$th-order between inputs from $m$ views \cite{cao2017deepmood}.
\begin{equation}
\begin{aligned}
\mathbf{q}_a^{(p)} &= \mathbf{U}_a^{(p)}[\mathbf{h}^{(p)}; 1] \\
\hat{y}_a &= \text{sum}([\mathbf{q}_a^{(1)}\odot\cdots\odot\mathbf{q}_a^{(m)}])
\end{aligned}
\label{eq:mvm}
\end{equation}
where $\mathbf{U}_a^{(p)} \in \mathbb{R}^{k \times (d_h+1)}$ is the factor matrix of the $p$-th view for the $a$-th class. By denoting $\bar{\mathbf{h}}^{(p)} = [\mathbf{h}^{(p)}; 1], p = 1, \cdots, m$, we can verify that Eq.~(\ref{eq:mvm}) is equivalent to Multi-view Machines \cite{cao2016multi}.

For experiments, the implementation is completed using Keras \cite{chollet2015keras} with Tensorflow \cite{tensorflow2015-whitepaper} as the backend. Features of alphanumeric characters, special characters and accelerometer values in a phone usage session are utilized to predict users' mood score. It is shown that the late fusion based DeepMood methods can achieve up to 90.31\% accuracy on predicting the depression score. It demonstrates the feasibility of using passive typing dynamics from mobile phone metadata to predict the disturbance and severity of mood states. In addition, it is found that the conventional shallow models like Support Vector Machine and Logistic Regression are not a good fit to this task, or sequence prediction in general. The tree boosting system \texttt{XGBoost}\footnote{https://github.com/dmlc/xgboost} \cite{chen2016xgboost} performs reasonably well as an ensemble method, but DeepMood still outperforms it by a significant margin 5.56\% in terms of prediction accuracy.

\begin{figure}[htb]
\centering
\includegraphics[width=3.3in]{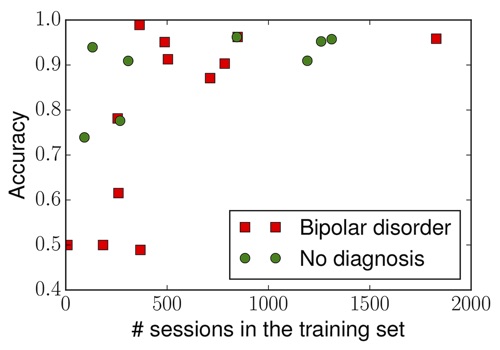}
\caption{Prediction performance on each of the 20 participants \cite{cao2017deepmood}.} \label{fig:mood}
\end{figure}

In practice, it is important to understand how the model works for each individual when monitoring her mood states. Therefore, we investigate the prediction performance on each of the 20 participants in our dataset. Results are shown in Fig. \ref{fig:mood} where each dot represents a participant with the number of her contributed sessions in the training set and the corresponding prediction accuracy. We can see that the proposed model can steadily produce accurate predictions ($\ge 87\%$) of a participant's mood states when she provides more than 400 valid typing sessions in the training phase. Note that the prediction we make in this work is per session which is typically less than one minute. We can expect more accurate results on the daily level by ensembling sessions occurring during a day.

The empirical analysis suggests that mobile device metadata could be used to predict the presence of mood disorders. The ability to passively collect data that can be used to infer the presence and severity of mood disturbances may enable providers to provide interventions to more patients earlier in their mood episodes, and it may also lead to deeper understanding of the effects of mood disturbances in the daily activities of people with mood disorders.

\subsection{User Identification}
User identification is a fundamental, but yet an open problem in mobile computing. Traditional approaches resort to user account information or browsing history. However, such information can pose security and privacy risks, and it is not robust as can be easily changed, e.g., the user changes to a new device or using a different application. Monitoring biometric information including a user's typing behaviors tends to produce consistent results over time while being less disruptive to user's experience. Furthermore, there are different kinds of sensors on mobile devices, meaning rich biometric information of users can be simultaneously collected. Thus, monitoring biometric information appears to be quite promising for mobile user identification.

In \cite{sun2017sequential}, we collect information from basic keystroke and the accelerometer on the phone, and then propose \textsc{DeepService}, a multi-view deep learning method, to utilize this information. To the best of our knowledge, this is the first time multi-view deep learning is applied to mobile user identification.

\textsc{DeepService} is a multi-view and multi-class identification framework via a deep structure. It contains three main steps to identify each user from several users.

\begin{itemize}
  \item In the first step, sequential tapping information and accelerometer information from volunteers who have used our provided mobile devices are collected. The sequential data is retrieved in a real-time manner.
  \item In the second step, the collected information is prepared as multi-view data for the problem of use identification.
  \item In the third step, the multi-view data is modeled via a deep structure to perform multi-class learning.
  \item In the last step, the performance of the proposed approach is compared with the traditional machine learning techniques for multi-class identification such as support vector machine and random forest.
\end{itemize}

\begin{figure}[tb]
\centering
\includegraphics[width=3.45in]{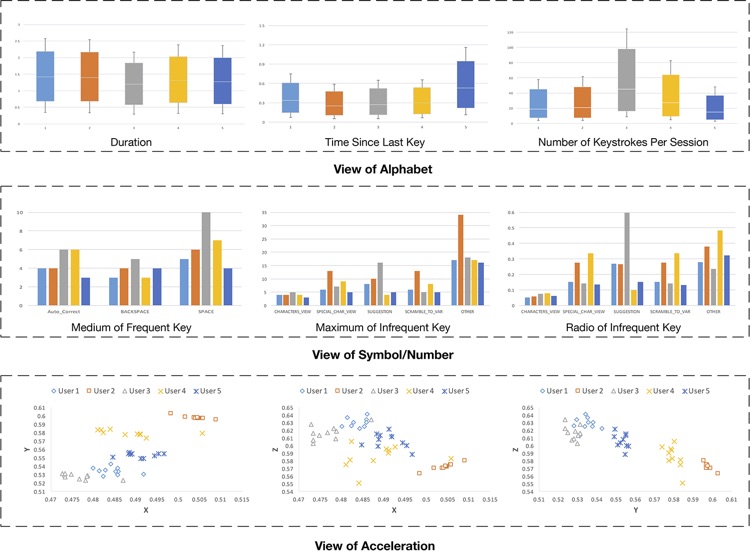}
\caption{Multi-view Pattern Analysis of Top 5 Active User: Left is \emph{user1} and Right is \emph{user5} \cite{sun2017sequential}} \label{fig:user_stat}
\end{figure}

The feature patterns for different users are evaluated. Fig.~\ref{fig:user_stat} shows the feature patterns analysis of top 5 active users in multi-view. 

In the view of Alphabet graphs, each user tends to have unique patterns with respect to the duration, the time since last key, and the number of keystrokes in each session. For example, \emph{user3} prefers to use more keystroke in every session with quicker tapping speed than other users.

In the view of Symbol/Number graphs, we have 8 different features. We separate these features into two groups: frequent keys and infrequent keys. A frequent key is defined as a key that is used more than twice per session, otherwise the key is an infrequent key. We show the medium number of keystroke per session of frequent keys such as auto correct, backspace and space. We also show the range and radio of infrequent keys per session of the top five active users. For example, \emph{user4} frequently uses auto\_correct, but she infrequently uses backspace.

In view of Acceleration graphs, we show the correlation of different directions of acceleration. From the last graph, we find that the top 5 active users can be well separated, which proves acceleration can help to identify the users well.

Our experimental results show that \textsc{DeepService} can do well identification between any two users with 98.97\% f1 score and 99.1\% accuracy in average. For example, private smartphone usually would not be shared by different people. However, sometimes the private phone could be shared between two people, such as husband and wife. \textsc{DeepService} can well separate any two people in this case.

For more general scenarios, we expand binary-identification to $N$ active class(user) identification to figure out who is using the mobile device either locally or on a web browser. Table~\ref{table:MVMC} reports our results.

\begin{table}[tb]
\caption{Results of \textsc{DeepService} and Baselines}
\centering
\label{table:MVMC}
\begin{tabular}{|l|c|c|c|c|}
\hline
                     & \multicolumn{2}{c|}{10}             & \multicolumn{2}{c|}{26}             \\ \hline
Method               &  Accuracy         & F1               & Accuracy         & F1               \\ \hline
LR             &  44.25\%          & 45.31\%          & 27.44\%          & 30.26\%          \\ \hline
SVM                  & 44.39\%          & 45.12\%          & 30.33\%          & 31.90\%          \\ \hline
Decision Tree        &  53.50\%          & 52.85\%          & 43.37\%          & 42.42\%          \\ \hline
RandomForest         &  77.05\%          & 76.59\%          & 67.87\%          & 66.31\%          \\ \hline
XGBoost              & 85.14\%          & 84.93\%          & 79.48\%          & 78.81\%          \\ \hline
\textbf{\sc DeepService} & \textbf{87.35\%} & \textbf{87.69\%} & \textbf{82.73\%} & \textbf{83.25\%} \\ \hline
\end{tabular}
\end{table}

If we increase the total number of the users in our model, it means we want to identify more people at the same time. For example, if our model is used on a home-router, it may need to identify only members of a family (3 to 10 people) at once. If we, instead, want to identify people working in a small office, we may need to identify more than 10 users. However, it is possible that a larger number of users would degrade the average performance of user identification. This is due to more variation of shared biometric patterns that introduce ambiguity into the system.

\section{Conclusion}
Pushing deep learning towards mobile devices has become a hot topic for both the academia and the industry. With the increasing demands of intelligent services on mobile devices, it is supposed to see continued leaps in deep learning based mobile applications in the next few years. Nonetheless, the current research about deep learning on mobile devices is just at the beginning. Encountered by the contradiction between high resource requirement of DNN and limited capacity of mobile devices, it is a challenging work to enabling deep learning services on mobile devices.

In this paper, we introduce the representative works about deep learning on mobile devices from three aspects: training by using mobile data, inference on mobile devices, and applications of mobile deep learning. We point out the main difficulties and solutions of these problems. Besides, we introduce our latest researches about deep learning on mobile devices. The summary of current works is provided in the paper. 

There are many unsolved and interesting problems about pushing deep learning towards mobile applications. Looking toward the next few years, with the rapid progress of mobile machine-learning processors and the emerge of next-generation wireless communication, mobile devices will be endowed with much more powerful computation capacity. We are likely to see more training tasks not just inference tasks being performed on mobile devices and growing numbers of mobile applications using deep learning being published by app providers and deployed by mobile users. More and more deep-learning applications will go out of the clouds and into the mobile devices, whether a smart phone or a embedded sensor. Nonetheless, the DNN's great demand for data would make the concerns about privacy and security of individuals' information mount, which may become a critical problem hindering the development of deep learning in mobile applications. How to overcome the two challenges about efficiency and privacy is supposed to be the main topic in this area. In addition, how to employ deep learning in diverse mobile applications is an important problem. Currently, deep learning is mainly applied to image and voice processing on mobile devices. Considering the versatility of mobile devices, we should attempt to empower more mobile applications by deep learning to make people's daily life more convenient and intelligent.

\bibliographystyle{IEEEtran}
\bibliography{mybib}

\begin{thebibliography}{10}
\providecommand{\url}[1]{#1}
\csname url@samestyle\endcsname
\providecommand{\newblock}{\relax}
\providecommand{\bibinfo}[2]{#2}
\providecommand{\BIBentrySTDinterwordspacing}{\spaceskip=0pt\relax}
\providecommand{\BIBentryALTinterwordstretchfactor}{4}
\providecommand{\BIBentryALTinterwordspacing}{\spaceskip=\fontdimen2\font plus
\BIBentryALTinterwordstretchfactor\fontdimen3\font minus
  \fontdimen4\font\relax}
\providecommand{\BIBforeignlanguage}[2]{{%
\expandafter\ifx\csname l@#1\endcsname\relax
\typeout{** WARNING: IEEEtran.bst: No hyphenation pattern has been}%
\typeout{** loaded for the language `#1'. Using the pattern for}%
\typeout{** the default language instead.}%
\else
\language=\csname l@#1\endcsname
\fi
#2}}
\providecommand{\BIBdecl}{\relax}
\BIBdecl

\bibitem{Cisco2016}
Cisco, ``Cisco visual networking index: forecast and methodology, 2015\-2020,''
  \emph{White Paper}, 2016.

\bibitem{He2017}
K.~He, G.~Gkioxari, P.~Doll{\'a}r, and R.~Girshick, ``Mask r-cnn,'' in
  \emph{Proceedings of the IEEE International Conference on Computer Vision
  (ICCV)}, 2017.

\bibitem{cao2017deepmood}
B.~Cao, L.~Zheng, C.~Zhang, P.~S. Yu, A.~Piscitello, J.~Zulueta, O.~Ajilore,
  K.~Ryan, and A.~D. Leow, ``{DeepMood}: Modeling mobile phone typing dynamics
  for mood detection,'' in \emph{KDD}.\hskip 1em plus 0.5em minus 0.4em\relax
  ACM, 2017, pp. 747--755.

\bibitem{Li2017modeling}
J.~Li, D.~Xiong, Z.~Tu, M.~Zhu, M.~Zhang, and G.~Zhou, ``Modeling source syntax
  for neural machine translation,'' in \emph{55th annual meeting of the
  Association for Computational Linguistics (ACL)}, 2017, pp. 4594--4602.

\bibitem{sun2016sigpid}
L.~Sun, Z.~Li, Q.~Yan, W.~Srisa-an, and Y.~Pan, ``Sigpid: significant
  permission identification for android malware detection,'' in \emph{11th
  International Conference on Malicious and Unwanted Software (MALWARE)}, 2016,
  pp. 1--8.

\bibitem{li2016droidclassifier}
Z.~Li, L.~Sun, Q.~Yan, W.~Srisa-an, and Z.~Chen, ``Droidclassifier: efficient
  adaptive mining of application-layer header for classifying android
  malware,'' in \emph{International Conference on Security and Privacy in
  Communication Systems (SecureComm)}, 2016, pp. 597--616.

\bibitem{Delloitte2017}
P.~Lee, ``Technology, media and telecommunications predictions,''
  \emph{Delloitte Touche Tohmatsu Limited}, 2017.

\bibitem{Goodfellow2016DL}
I.~Goodfellow, Y.~Bengio, and A.~Courville, \emph{Deep Learning}.\hskip 1em
  plus 0.5em minus 0.4em\relax Cambridge, MA, USA: The MIT Press, 2016.

\bibitem{Lane2017}
N.~D. Lane, S.~Bhattacharya, A.~Mathur, P.~Georgiev, C.~Forlivesi, and
  F.~Kawsar, ``Squeezing deep learning into mobile and embedded devices,''
  \emph{IEEE Pervasive Computing}, vol.~16, no.~3, pp. 82--88, 2017.

\bibitem{Kingm2015}
D.~P. Kingma and J.~Ba, ``Adam: A method for stochastic optimization,'' in
  \emph{International Conference on Learning Representations (ICLR)}, 2015.

\bibitem{Duchi2011}
J.~Duchi, E.~Hazan, and Y.~Singer, ``Adaptive subgradient methods for online
  learning and stochastic optimization,'' \emph{Journal of Machine Learning
  Research}, vol.~12, no.~6, pp. 2121--2159, 2011.

\bibitem{Rmsprop}
T.~Tieleman and G.~E. Hinto, ``Lectur e6.5-rmsprop: Divide the gradient by a
  running average of its recent magnitude,'' \emph{COURSERA: Neural Networks
  for Machine Learnings}, vol.~4, no.~2, 2012.

\bibitem{Han2015learning}
S.~Han, J.~Pool, J.~Tran, and W.~J. Dally, ``Learning both weights and
  connections for efficient neural networks,'' in \emph{Proceedings of the 28th
  International Conference on Neural Information Processing Systems (NIPS)},
  2015, pp. 1135--1143.

\bibitem{Ding2017}
C.~Ding, S.~Liao, Y.~Wang, Z.~Li, N.~Liu, Y.~Zhuo, C.~Wang, X.~Qian, Y.~Bai,
  G.~Yuan, X.~Ma, Y.~Zhang, J.~Tang, Q.~Qiu, X.~Lin, and B.~Yuan, ``Circnn:
  Accelerating and compressing deep neural networks using block-circulant
  weight matrices,'' in \emph{Proceedings of the 50th Annual IEEE/ACM
  International Symposium on Microarchitecture (MICRO)}.\hskip 1em plus 0.5em
  minus 0.4em\relax ACM, 2017, pp. 395--408.

\bibitem{Avriel2003}
M.~Avriel, \emph{Nonlinear Programming: Analysis and Methods}.\hskip 1em plus
  0.5em minus 0.4em\relax Dover Publications, 2003.

\bibitem{Shokri2015}
R.~Shokri and V.~Shmatikov, ``Privacy-preserving deep learning,'' in
  \emph{Proceedings of the 22nd ACM SIGSAC Conference on Computer and
  Communications Security (CCS)}, 2015, pp. 1310--1321.

\bibitem{McMahan2017}
B.~McMahan and D.~Ramage, ``Federated learning: Collaborative machine learning
  without centralized training data,'' \emph{Google Research Blog}, 2017.

\bibitem{McMahan2016}
H.~B. McMahan, E.~Moore, and D.~R. D.~A. y~Arcas, ``Federated learning of deep
  networks using model averaging,'' \emph{arXiv:1602.05629}, 2016.

\bibitem{Hitaj2017}
B.~Hitaj, G.~Ateniese, and F.~Perez-Cruz, ``Deep models under the gan:
  Information leakage from collaborative deep learning,'' in \emph{Proceedings
  of ACM SIGSAC Conference on Computer and Communications Security
  (CCS)}.\hskip 1em plus 0.5em minus 0.4em\relax ACM, 2017, pp. 603--618.

\bibitem{Abadi2016}
M.~Abadi, A.~Chu, I.~Goodfellow, H.~B. McMahan, I.~Mironov, K.~Talwar, and
  L.~Zhang, ``Deep learning with differential privacy,'' in \emph{Proceedings
  of the 23rd ACM SIGSAC Conference on Computer and Communications Security
  (CCS)}, 2016, pp. 308--318.

\bibitem{Papernot2017}
N.~Papernot, M.~Abadi, U.~Erlingsson, I.~Goodfellow, and K.~Talwar,
  ``Semi-supervised knowledge transfer for deep learning from private training
  data,'' in \emph{5th International Conference on Learning Representations
  (ICLR)}, 2017.

\bibitem{McMahan2017learning}
H.~B. McMahan, D.~Ramage, K.~Talwar, and L.~Zhang, ``Learning differentially
  private recurrent language models,'' \emph{arXiv:1710.06963}, 2017.

\bibitem{Beimel2014}
A.~Beimel, H.~Brenner, S.~P. Kasiviswanathan, and K.~Nissim, ``Bounds on the
  sample complexity for private learning and private data release,''
  \emph{Machine Learning}, vol.~94, no.~3, pp. 401--437, 2014.

\bibitem{Dwork2011diff}
C.~Dwork, \emph{Differential Privacy}.\hskip 1em plus 0.5em minus 0.4em\relax
  Boston, MA: Springer US, 2011, pp. 338--340.

\bibitem{Teerapittayanon2017}
S.~Teerapittayanon, B.~McDanel, and H.~T. Kung, ``Distributed deep neural
  networks over the cloud, the edge and end devices,'' in \emph{IEEE 37th
  International Conference on Distributed Computing Systems (ICDCS)}, 2017, pp.
  328--339.

\bibitem{Osia2017}
S.~A. Osia, A.~S. Shamsabadi, A.~Taheri, H.~R. Rabiee, N.~D. Lane, and
  H.~Haddadi, ``A hybrid deep learning architecture for privacy-preserving
  mobile analytics,'' \emph{arXiv:1703.02952}, 2017.

\bibitem{Li2017}
M.~Li, L.~Lai, N.~Suda, V.~Chandra, and D.~Z. Pan, ``Privynet: A flexible
  framework for privacy-preserving deep neural network training with a
  fine-grained privacy control,'' \emph{arXiv:1709.06161}, 2017.

\bibitem{Han2016}
S.~Han, H.~Mao, and W.~J. Dally, ``Deep compression: Compressing deep neural
  networks with pruning, trained quantization and huffman coding,'' in
  \emph{4th International Conference on Learning Representations (ICLR)}, 2016.

\bibitem{Howard2017}
A.~G. Howard, M.~Zhu, B.~Chen, D.~Kalenichenko, W.~Wang, T.~Weyand,
  M.~Andreetto, and H.~Adam, ``Mobilenets: Efficient convolutional neural
  networks for mobile vision applications,'' \emph{arXiv:1704.04861}, 2017.

\bibitem{wang2018not}
J.~Wang, J.~Zhang, W.~Bao, X.~Zhu, B.~Cao, and P.~S. Yu, ``Not just privacy:
  Improving performance of private deep learning in mobile cloud,'' in
  \emph{Proceedings of the 24th ACM SIGKDD International Conference on
  Knowledge Discovery \& Data Mining (KDD)}, 2018, pp. 2407--2416.

\bibitem{Cheng2017}
Y.~Cheng, D.~Wang, P.~Zhou, and T.~Zhang, ``A survey of model compression and
  acceleration for deep neural networks,'' \emph{IEEE Signal Processing
  Magazine}, vol. online, 2017.

\bibitem{Gong2014}
Y.~Gong, L.~Liu, M.~Yang, and L.~Bourdev, ``Compressing deep convolutional
  networks using vector quantization,'' \emph{arXiv:1412.6115}, 2014.

\bibitem{Wu2016}
J.~Wu, C.~Leng, Y.~Wang, Q.~Hu, and J.~Cheng, ``Quantized convolutional neural
  networks for mobile devices,'' in \emph{IEEE Conference on Computer Vision
  and Pattern Recognition (CVPR)}, 2016.

\bibitem{Gupta2015}
S.~Gupta, A.~Agrawal, K.~Gopalakrishnan, and P.~Narayanan, ``Deep learning with
  limited numerical precision,'' in \emph{International Conference on Learning
  Representations (ICLR)}, 2016.

\bibitem{Cheng2015an}
Y.~Cheng, F.~X. Yu, R.~S. Feris, S.~Kumar, A.~Choudhary, and S.-F. Chang, ``An
  exploration of parameter redundancy in deep networks with circulant
  projections,'' in \emph{Proceedings of the IEEE International Conference on
  Computer Vision (ICCV)}, 2015.

\bibitem{Denton2014}
E.~L. Denton, W.~Zaremba, J.~Bruna, Y.~LeCun, and R.~Fergus, ``Exploiting
  linear structure within convolutional networks for efficient evaluation,'' in
  \emph{Advances in Neural Information Processing Systems (NIPS)}, 2014, pp.
  1269--1277.

\bibitem{Hinton2015}
G.~Hinton, O.~Vinyals, and J.~Dean, ``Distilling the knowledge in a neural
  network,'' in \emph{Advances in Neural Information Processing Systems
  (NIPS)}, 2014.

\bibitem{ankers2009objective}
D.~Ankers and S.~H. Jones, ``Objective assessment of circadian activity and
  sleep patterns in individuals at behavioural risk of hypomania,''
  \emph{Journal of clinical psychology}, vol.~65, no.~10, pp. 1071--1086, 2009.

\bibitem{bopp2010longitudinal}
J.~M. Bopp, D.~J. Miklowitz, G.~M. Goodwin, W.~Stevens, J.~M. Rendell, and
  J.~R. Geddes, ``The longitudinal course of bipolar disorder as revealed
  through weekly text messaging: a feasibility study,'' \emph{Bipolar
  disorders}, vol.~12, no.~3, pp. 327--334, 2010.

\bibitem{faurholt2016behavioral}
M.~Faurholt-Jepsen, M.~Vinberg, M.~Frost, S.~Debel, E.~Margrethe~Christensen,
  J.~E. Bardram, and L.~V. Kessing, ``Behavioral activities collected through
  smartphones and the association with illness activity in bipolar disorder,''
  \emph{International journal of methods in psychiatric research}, vol.~25,
  no.~4, pp. 309--323, 2016.

\bibitem{cho2014learning}
K.~Cho, B.~Van~Merri{\"e}nboer, C.~Gulcehre, D.~Bahdanau, F.~Bougares,
  H.~Schwenk, and Y.~Bengio, ``Learning phrase representations using rnn
  encoder-decoder for statistical machine translation,'' \emph{arXiv preprint
  arXiv:1406.1078}, 2014.

\bibitem{hochreiter1997long}
S.~Hochreiter and J.~Schmidhuber, ``Long short-term memory,'' \emph{Neural
  computation}, vol.~9, no.~8, pp. 1735--1780, 1997.

\bibitem{cao2016multi}
B.~Cao, H.~Zhou, G.~Li, and P.~S. Yu, ``Multi-view machines,'' in
  \emph{WSDM}.\hskip 1em plus 0.5em minus 0.4em\relax ACM, 2016, pp. 427--436.

\bibitem{rendle2012factorization}
S.~Rendle, ``Factorization machines with libfm,'' \emph{ACM Transactions on
  Intelligent Systems and Technology}, vol.~3, no.~3, p.~57, 2012.

\bibitem{chollet2015keras}
F.~Chollet, ``Keras,'' \url{https://github.com/fchollet/keras}, 2015.

\bibitem{tensorflow2015-whitepaper}
\BIBentryALTinterwordspacing
M.~A. et~al., ``{TensorFlow}: Large-scale machine learning on heterogeneous
  systems,'' 2015, software available from tensorflow.org. [Online]. Available:
  \url{http://tensorflow.org/}
\BIBentrySTDinterwordspacing

\bibitem{chen2016xgboost}
T.~Chen and C.~Guestrin, ``Xgboost: A scalable tree boosting system,'' in
  \emph{KDD}.\hskip 1em plus 0.5em minus 0.4em\relax ACM, 2016.

\bibitem{sun2017sequential}
L.~Sun, Y.~Wang, B.~Cao, S.~Y. Philip, W.~Srisa-an, and A.~D. Leow,
  ``Sequential keystroke behavioral biometrics for mobile user identification
  via multi-view deep learning,'' in \emph{Joint European Conference on Machine
  Learning and Knowledge Discovery in Databases}, 2017, pp. 228--240.

\end{thebibliography}

\end{document}